\documentclass[10pt,twocolumn,letterpaper]{article}

\usepackage{iccv}
\usepackage{times}
\usepackage{epsfig}
\usepackage{graphicx}
\usepackage{amsmath}
\usepackage{amssymb}
\usepackage{multirow}
\usepackage{subfigure}
\usepackage{verbatim}
\usepackage{colortbl}
\usepackage{xcolor}

\newcommand*\samethanks[1][\value{footnote}]{\footnotemark[#1]}

\usepackage[pagebackref=true,breaklinks=true,letterpaper=true,colorlinks,bookmarks=false]{hyperref}

\iccvfinalcopy

\def\iccvPaperID{4495}

\ificcvfinal\fi
\begin{document}

\title{High-Resolution Representations
for Labeling Pixels and Regions}

\author{Ke Sun$^{1,2}$$\thanks{Equal contribution.}$ ~~~ Yang Zhao$^{3}$\samethanks[1]~~~Borui Jiang$^{2,4}$\samethanks[1] ~~~ Tianheng Cheng$^{2,5}$\samethanks[1]
	~~~ Bin Xiao$^{2}$ \\ Dong Liu$^{1}$~~~ Yadong Mu$^{4}$~~~Xinggang Wang$^{5}$~~~Wenyu Liu$^{5}$~~~ Jingdong Wang$^{2}$$\thanks{Corresponding author, welleast@outlook.com}$\\
	$^{1}$University of Science and Technology of China~~~ $^{2}$Microsoft Research Asia\\$^{3}$The University of Adelaide $^{4}$Peking University $^{5}$Huazhong University of Science and Technology\\	
	{\tt\small sunk@mail.ustc.edu.cn, yang.zhao4@griffithuni.edu.au, jbr@pku.edu.cn, muyadong@gmail.com}\\ {\tt\small \{vic,liuwy,xgwang\}@hust.edu.cn, dongleiu@ustc.edu.cn, \{Bin.Xiao,jingdw\}@microsoft.com}
}

\maketitle
\thispagestyle{empty}
\begin{abstract}
High-resolution representation learning
plays an essential role in many vision problems,
e.g., pose estimation and semantic segmentation.
The high-resolution network (HRNet)~\cite{SunXLW19},
recently developed for human pose estimation,
maintains high-resolution representations
through the whole process
by connecting high-to-low resolution convolutions in \emph{parallel}
and produces strong high-resolution representations
by repeatedly conducting fusions across parallel convolutions.

In this paper, we conduct a further study
on high-resolution representations
by introducing a simple yet effective modification
and apply it to a wide range of vision tasks.
We augment the high-resolution representation
by aggregating the (upsampled) representations
from all the parallel convolutions
rather than only the representation
from the high-resolution convolution as done in~\cite{SunXLW19}.
This simple modification leads to stronger representations,
evidenced by superior results.
We show top results
in semantic segmentation
on Cityscapes, LIP,
and PASCAL Context,
and facial landmark detection on AFLW, COFW, $300$W, and WFLW.
In addition,
we build a multi-level representation
from the high-resolution representation
and apply it to the Faster R-CNN object detection framework and the extended frameworks.
The proposed approach achieves superior results
to existing single-model networks on COCO object detection.
The code and models have been publicly available at \url{https://github.com/HRNet}.
\end{abstract}

\section{Introduction}
Deeply-learned representations
have been demonstrated
to be strong and
achieved state-of-the-art results in many vision tasks.
There are two main kinds of representations:
low-resolution representations
that are mainly for image classification,
and high-resolution representations
that are essential for many other vision problems,
e.g., semantic segmentation,
object detection,
human pose estimation,
etc.
The latter one,
the interest of this paper,
remains unsolved and is attracting a lot of attention.

There are two main lines for
computing high-resolution representations.
One is to recover high-resolution representations from low-resolution representations outputted by a network
(e.g., ResNet)
and optionally intermediate medium-resolution representations,
e.g., Hourglass~\cite{NewellYD16},
SegNet~\cite{BadrinarayananK17},
DeconvNet~\cite{NohHH15},
U-Net~\cite{RonebergerFB15},
and encoder-decoder~\cite{PengFWM16}.
The other one is to maintain high-resolution representations
through high-resolution convolutions
and strengthen the representations
with parallel low-resolution convolutions~\cite{SunXLW19,FourureEFMT017,ZhouHZ15,SaxenaV16}.
In addition, dilated convolutions
are used to replace some strided convolutions
and associated regular convolutions
in classification networks
to compute medium-resolution representations~\cite{ChenPKMY18,ZhaoSQWJ17}.

We go along the research line
of maintaining high-resolution representations
and further study the high-resolution network (HRNet),
which is initially developed for human pose estimation~\cite{SunXLW19},
for a broad range of vision tasks.
An HRNet maintains high-resolution representations
by connecting high-to-low resolution convolutions in parallel
and repeatedly conducting multi-scale fusions across parallel convolutions.
The resulting high-resolution representations are not only strong
but also spatially precise.

\begin{figure*}[t]
\footnotesize
    \centering
    \includegraphics[width = 0.99\textwidth]{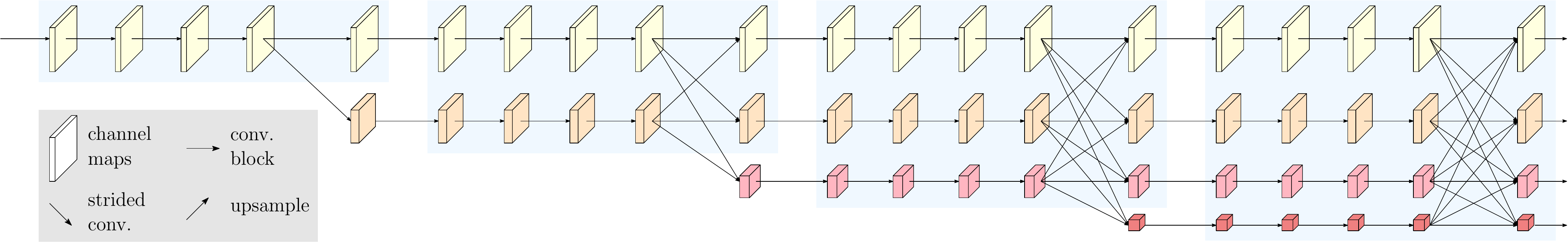}
    \caption{A simple example
    of a high-resolution network.
    There are four stages.
    The $1$st stage consists of high-resolution convolutions.
    The $2$nd ($3$rd, $4$th) stage
    repeats two-resolution
    (three-resolution, four-resolution)
    blocks. The detail is given in Section~\ref{sec:HRNetV1}.}
    \label{fig:HRNet}
    \vspace{-2mm}
\end{figure*}

We make a simple modification
by exploring the representations
from all the high-to-low resolution parallel convolutions
other than only the high-resolution representations in the original HRNet ~\cite{SunXLW19}.
This modification adds a small overhead and
leads to stronger high-resolution representations.
The resulting network is named as HRNetV$2$.
We empirically show the superiority to the original HRNet.

We apply our proposed network
to semantic segmentation/facial landmark detection
through estimating segmentation maps/facial landmark heatmaps
from the output high-resolution representations.
In semantic segmentation,
the proposed approach achieves
state-of-the-art results
on PASCAL Context, Cityscapes, and LIP
with similar model sizes and lower computation complexity.
In facial landmark detection,
our approach achieves overall best results
on four standard datasets: AFLW, COFW, $300$W, and WFLW.

In addition, we construct a multi-level representation
from the high-resolution representation,
and apply it to the Faster R-CNN object detection framework
and its extended frameworks, Mask R-CNN \cite{HeGDG17} and Cascade R-CNN \cite{CaiV18}.
The results show that our method gets great detection performance improvement and in particular dramatic improvement
for small objects.
With single-scale training and testing,
the proposed approach achieves better COCO object detection results
than existing single-model methods.

\section{Related Work}
Strong high-resolution representations play an essential role
in pixel and region labeling problems,
e.g., semantic segmentation, human pose estimation, facial landmark detection,
and object detection.
We review representation learning techniques developed
mainly in the semantic segmentation,
facial landmark detection~\cite{SunWT13,KowalskiNT17,LvSXCZ17,XiaoFXLYK16,ZhangLLT14,TrigeorgisSNAZ16,ZhangSKC14} and object detection areas\footnote{The techniques developed for human pose estimation
are reviewed in~\cite{SunXLW19}.},
from low-resolution representation learning,
high-resolution representation recovering,
to high-resolution representation maintaining.

\noindent\textbf{Learning low-resolution representations.}
The fully-convolutional network (FCN) approaches~\cite{LongSD15,SermanetEZMFL13}
compute low-resolution representations
by removing the fully-connected layers in a classification network,
and estimate from their coarse segmentation confidence maps.
The estimated segmentation maps are improved
by combining the fine segmentation score maps
estimated from intermediate low-level medium-resolution representations~\cite{LongSD15},
or iterating the processes~\cite{KowalskiNT17}.
Similar techniques have also been applied
to edge detection, e.g., holistic edge detection~\cite{XieT15}.

The fully convolutional network
is extended,
by replacing a few (typically two) strided convolutions
and the associated convolutions with dilated convolutions,
to the dilation version,
leading to medium-resolution representations~\cite{ZhaoSQWJ17,ChenPKMY18, YuKF17,ChenPKMY14, LiPYZDS18}.
The representations are further augmented
to multi-scale contextual representations~\cite{ZhaoSQWJ17,ChenPKMY18,ChenYWXY16} through feature pyramids
for segmenting objects at multiple scales.

\vspace{.1cm}
\noindent\textbf{Recovering high-resolution representations.}
An upsample subnetwork,
like a decoder,
is adopted
to gradually recover the high-resolution representations
from the low-resolution representations outputted by the downsample process.
The upsample subnetwork could be a symmetric version
of the downsample subnetwork,
with skipping connection over some mirrored layers
to transform the pooling indices,
e.g., SegNet~\cite{BadrinarayananK17} and DeconvNet~\cite{NohHH15},
or copying the feature maps, e.g., U-Net~\cite{RonebergerFB15} and Hourglass~\cite{NewellYD16, YangLZ17, BulatT17, DengTZZ17, BulatT17a},
encoder-decoder~\cite{PengFWM16},
FPN~\cite{LinDGHHB17}, and so on.
The full-resolution residual network~\cite{PohlenHML17}
introduces an extra full-resolution stream
that carries information at the full image resolution,
to replace the skip connections,
and each unit in the downsample and upsample subnetworks
receives information from and sends information to
the full-resolution stream.

The asymmetric upsample process is also widely studied.
RefineNet~\cite{LinMSR17} improves the combination
of upsampled representations and
the representations of the same resolution
copied from the downsample process.
Other works include:
light upsample process~\cite{BulatT16};
light downsample and heavy upsample processes~\cite{ValleBVB18},
recombinator networks~\cite{HonariYVP16};
improving skip connections with more or complicated convolutional units~\cite{PengZYLS17, ZhangZPXS18, IslamRBW17}, as well as sending information from low-resolution skip connections to high-resolution skip connections~\cite{ZhouSTL18}
or exchanging information between them~\cite{GuoDXZ18};
studying the details the upsample process~\cite{WojnaUGSCFF17};
combining multi-scale pyramid representations~\cite{ChenZPSA18, XiaoLZJS18};
stacking multiple DeconvNets/U-Nets/Hourglass~\cite{FuLWL17, Wu0YWC018}
with dense connections~\cite{TangPGWZM18}.

\vspace{.1cm}
\noindent\textbf{Maintaining high-resolution representations.}
High-resolution representations
are maintained through the whole process,
typically by a network that is formed
by connecting multi-resolution (from high-resolution to low-resolution) parallel convolutions
with repeated information exchange across parallel convolutions.
Representative works include GridNet~\cite{FourureEFMT017}, convolutional neural fabrics~\cite{SaxenaV16}, interlinked CNNs~\cite{ZhouHZ15},
and the recently-developed high-resolution networks (HRNet)~\cite{SunXLW19}
that is our interest.

The two early works, convolutional neural fabrics~\cite{SaxenaV16}
and interlinked CNNs~\cite{ZhouHZ15},
lack careful design
on when to start low-resolution parallel streams
and how and when to exchange information across parallel streams,
and do not use batch normalization and residual connections,
thus not showing satisfactory performance.

GridNet~\cite{FourureEFMT017}
is like a combination of multiple U-Nets
and includes two symmetric information exchange stages:
the first stage only passes information from high-resolution
to low-resolution,
and the second stage only passes information
from low-resolution to high-resolution.
This limits its segmentation quality.

\begin{figure}[t]
    \centering
    \footnotesize
   \subfigure[]{\includegraphics[height=0.33\linewidth]{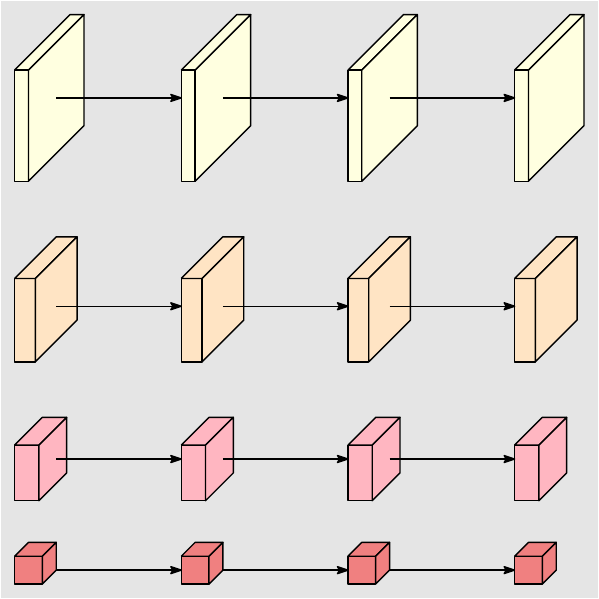}}~~
    \subfigure[]{\includegraphics[height=0.33\linewidth]{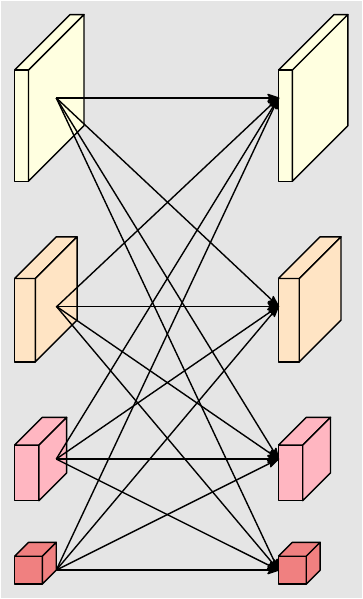}}~~
    \subfigure[]{\includegraphics[height=0.33\linewidth]{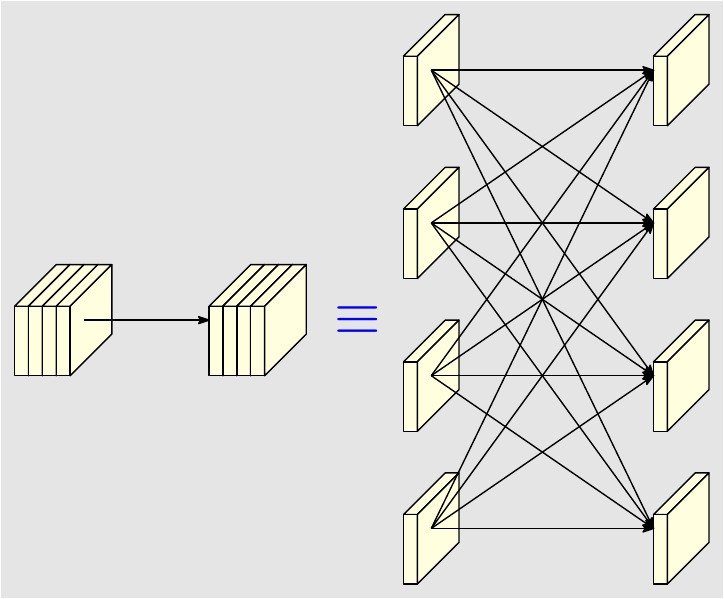}}
    \caption{Multi-resolution block:
    (a) multi-resolution group convolution and (b) multi-resolution convolution.
    (c) A normal convolution (left) is equivalent to
    fully-connected multi-branch convolutions (right).}
    \label{fig:multiresolutionblock}
    \vspace{-3mm}
\end{figure}

\begin{figure*}[t]
\footnotesize
    \centering
    (a)~~\includegraphics[scale=.8,angle=90]{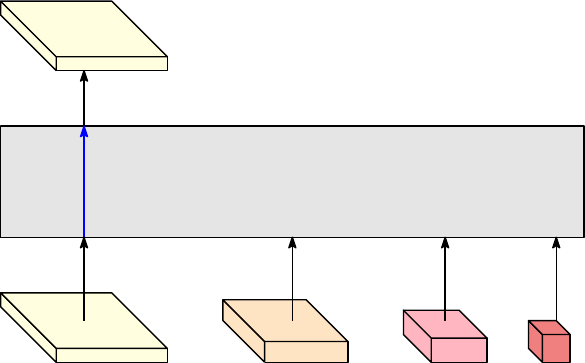}~~~~~~~~~~
    (b)~~\includegraphics[scale=.8,angle=90]{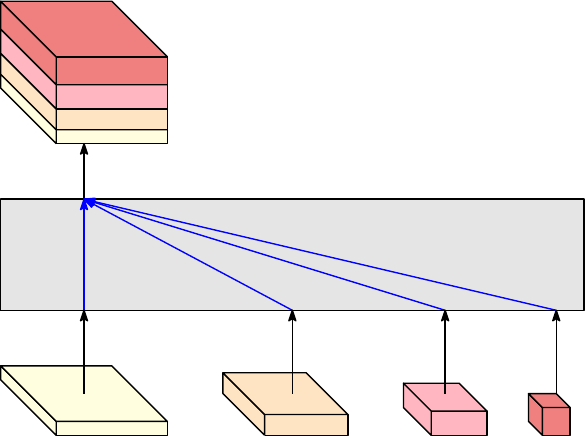}~~~~~~~~~~
    (c)~~\includegraphics[scale=.8,angle=90]{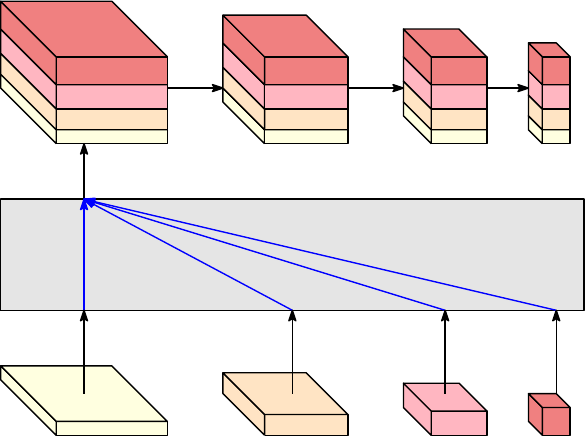}
    \caption{
    (a) The high-resolution representation proposed in~\cite{SunXLW19} (HRNetV$1$ );
    (b) Concatenating the (upsampled) representations
    that are from all the resolutions
    for semantic segmentation and facial landmark detection (HRNetV$2$ );
    (c) A feature pyramid formed over (b)
    for object detection (HRNetV$2$p).
    The four-resolution representations at the bottom in each sub-figure are outputted from the network in Figure~\ref{fig:HRNet},
    and the gray box indicates how the output representation is obtained from the input four-resolution representations.}
    \label{fig:highresolutionhead}
    \vspace{-3mm}
\end{figure*}

\section{Learning High-Resolution Representations}
\label{sec:HRNetV1}
The high-resolution network~\cite{SunXLW19}, which we named HRNetV$1$
for convenience,
maintains high-resolution representations
by connecting high-to-low resolution convolutions in parallel,
where there are repeated multi-scale fusions across parallel convolutions.

\vspace{.1cm}
\noindent\textbf{Architecture.}
The architecture is illustrated in Figure~\ref{fig:HRNet}.
There are four stages,
and the $2$nd, $3$rd and $4$th stages are formed
by repeating modularized multi-resolution blocks.
A multi-resolution block
consists of a multi-resolution group convolution
and a multi-resolution convolution
which is illustrated in Figure~\ref{fig:multiresolutionblock} (a) and (b).
The multi-resolution group convolution is a simple extension
of the group convolution, which divides the input channels
into several subsets of channels
and performs a regular convolution
over each subset
over different spatial resolutions separately.

The multi-resolution convolution is depicted
in Figure~\ref{fig:multiresolutionblock} (b).
It resembles the multi-branch full-connection manner
of the regular convolution, illustrated in in Figure~\ref{fig:multiresolutionblock} (c).
A regular convolution
can be divided as multiple small convolutions
as explained in~\cite{ZhangQ0W17}.
The input channels are divided into several subsets,
and the output channels are also divided into several subsets.
The input and output subsets are connected
in a fully-connected fashion,
and each connection is a regular convolution.
Each subset of output channels
is a summation of the outputs of the convolutions
over each subset of input channels.

The differences lie in two-fold.
(i) In a multi-resolution convolution
each subset of channels is over a different resolution.
(ii)
The connection between input channels
and output channels
needs to handle
The resolution decrease is
implemented in~\cite{SunXLW19}
by using several $2$-strided $3\times 3$ convolutions.
The resolution increase is simply implemented in~\cite{SunXLW19}
by bilinear (nearest neighbor) upsampling.

\vspace{.1cm}
\noindent\textbf{Modification.}
In the original approach HRNetV$1$,
only the representation (feature maps)
from the high-resolution convolutions
in~\cite{SunXLW19}
are outputted,
which is illustrated in Figure~\ref{fig:highresolutionhead} (a).
This means that
only a subset of output channels
from the high-resolution convolutions
is exploited
and other subsets from low-resolution convolutions are lost.

We make a simple yet effective modification
by exploiting other subsets of channels
outputted from low-resolution convolutions.
The benefit is that the capacity of the multi-resolution
convolution is fully explored.
This modification only adds a small parameter and computation overhead.

We rescale the low-resolution representations through bilinear upsampling
to the high resolution,
and concatenate the subsets of representations,
illustrated in Figure~\ref{fig:highresolutionhead} (b),
resulting in the high-resolution representation,
which we adopt for estimating segmentation maps/facial landmark heatmaps.
In application to object detection,
we construct a multi-level representation
by downsampling the high-resolution representation
with average pooling
to multiple levels,
which is depicted in Figure~\ref{fig:highresolutionhead} (c).
We name the two modifications as HRNetV$2$ and HRNetV$2$p, respectively,
and empirically compare them in Section~\ref{sec:empiricalstudy}.

\vspace{.1cm}
\noindent\textbf{Instantiation}
We instantiate the network using a similar manner as HRNetV$1$~\cite{SunXLW19}\footnote{\url{https://github.com/leoxiaobin/deep-high-resolution-net.pytorch}}.
The network starts from a stem
that consists of two strided $3 \times 3$ convolutions
decreasing the resolution to $1/4$.
The $1$st stage contains $4$ residual units
where each unit
is formed by a bottleneck with the width $64$,
and is followed by one $3\times3$ convolution
reducing the width of feature maps
to $C$.
The $2$nd, $3$rd, $4$th stages
contain $1$, $4$, $3$ multi-resolution blocks, respectively.
The widths (number of channels) of the convolutions
of the four resolutions
are $C$, $2C$, $4C$, and $8C$, respectively.
Each branch in the multi-resolution group convolution
contains $4$ residual units
and each unit contains two $3\times3$ convolutions in each resolution.

In applications to semantic segmentation
and facial landmark detection,
we mix the output representations (Figure~\ref{fig:highresolutionhead} (b)),
from all the four resolutions
through a $1 \times 1$ convolution,
and produce a $15C$-dimensional representation.
Then, we pass the mixed representation
at each position
to a linear classifier/regressor
with the softmax/MSE loss
to predict the segmentation maps/facial landmark heatmaps.
For semantic segmentation,
the segmentation maps are upsampled ($4$ times)
to the input size by bilinear upsampling
for both training and testing.
In application to object detection,
we reduce the dimension of the high-resolution representation
to $256$, similar to FPN~\cite{LinDGHHB17},
through a $1\times 1$ convolution
before forming the feature pyramid in Figure~\ref{fig:highresolutionhead} (c).

\section{Experiments}
\subsection{Semantic Segmentation}
Semantic segmentation
is a problem of assigning
a class label to each pixel.
We report the results over
two scene parsing datasets, PASCAL Context~\cite{MottaghiCLCLFUY14} and
Cityscapes~\cite{CordtsORREBFRS16},
and a human parsing dataset, LIP~\cite{GongLSL17}. The mean of class-wise intersection over union (mIoU)
is adopted as the evaluation metric.

	\setlength{\tabcolsep}{6.0pt}
	\begin{table}[t]
		\scriptsize
		\centering
		\caption{Segmentation results on
		Cityscapes \texttt{val}
	    (single scale and no flipping).
		The GFLOPs is calculated on the input size $1024 \times 2048$.}
		\label{tab:cityscapevalresults}
		\begin{tabular}{l|lrr|c}
			\hline\noalign{\smallskip}
			 & backbone & \#param. & GFLOPs & mIoU\\
			 \hline
			
			 \hline
			UNet++~\cite{ZhouSTL18} & ResNet-$101$ & $59.5$M & $748.5$ & $75.5$ \\
			DeepLabv3~\cite{ChenPSA17} & Dilated-ResNet-$101$ & $58.0$M & $1778.7$ & $78.5$ \\
			DeepLabv3+~\cite{ChenZPSA18} & Dilated-Xception-$71$ & $43.5$M & $1444.6$ & $79.6$ \\
			PSPNet~\cite{ZhaoSQWJ17} & Dilated-ResNet-$101$ & $65.9$M & $2017.6$ & $79.7$ \\
			\hline
			Our approach &  HRNetV$2$-W$40$ & $45.2$M & $493.2$ & $80.2$ \\
			Our approach & HRNetV$2$-W$48$ & $65.9$M & $747.3$ & $\mathbf{81.1}$ \\
			\hline
		\end{tabular}
		\vspace{-4mm}
	\end{table}

\vspace{.1cm}
\noindent\textbf{Cityscapes.}
The Cityscapes dataset~\cite{CordtsORREBFRS16} contains $5,000$ high quality pixel-level finely annotated scene images.
The finely-annotated images are divided into $2,975/500/1,525$ images for training, validation and testing.
There are $30$ classes, and $19$ classes among them are used for evaluation.
In addition to the mean of class-wise intersection over union (mIoU),
we report other three scores on the test set:
IoU category (cat.), iIoU class (cla.) and iIoU category (cat.).

We follow the same training protocol~\cite{ZhaoSQWJ17, ZhaoZLSLLJ18}.
The data are augmented by random cropping (from $1024\times2048$ to $512\times1024$), random scaling in the range of $[0.5, 2]$, and random horizontal flipping. We use the SGD optimizer with the base learning rate of $0.01$,
the momentum of $0.9$ and the weight decay of $0.0005$. The poly learning rate policy with the power of $0.9$ is used for dropping the learning rate. All the models are trained for $120K$ iterations with the batch size of $12$ on $4$ GPUs and syncBN.

\renewcommand{\arraystretch}{1.3}
	\setlength{\tabcolsep}{3pt}
	\begin{table}[t]
		\scriptsize
		\centering
		\caption{Semantic segmentation results on Cityscapes \texttt{test}.}
		\label{tab:cityscaperesults}
		\begin{tabular}{l|l|cccc}
			\hline\noalign{\smallskip}
			  & backbone & mIoU  & iIoU cla. & IoU cat. & iIoU cat.\\
			\hline
			
			\hline
			\multicolumn{3}{l}{\emph {Model learned on the \texttt{train} set}}\\
			\hline
			PSPNet~\cite{ZhaoSQWJ17} & Dilated-ResNet-$101$ & $78.4$ & $56.7$ & $90.6$  & $78.6$ \\
			PSANet~\cite{ZhaoZLSLLJ18} & Dilated-ResNet-$101$ & $78.6$ & - & - & - \\
			PAN~\cite{LiXAW18} & Dilated-ResNet-$101$ & $78.6$ & - & - & - \\
			AAF~\cite{KeHLY18} & Dilated-ResNet-$101$ & $79.1$ & - & - & -\\
			\hline
			Our approach & HRNetV$2$-W$48$ & $\mathbf{80.4}$ & $\mathbf{59.2}$ & $\mathbf{91.5}$ & $\mathbf{80.8}$\\
			\hline
			
			\hline
			\multicolumn{3}{l}{
			\emph {Model learned on the \texttt{train+valid} set}}\\
			\hline
			GridNet~\cite{FourureEFMT017} & - & $69.5$ & $44.1$ & $87.9$ & $71.1$\\
			LRR-4x~\cite{GhiasiF16} & - & $69.7$ & $48.0$ & $88.2$ & $74.7$\\
			DeepLab~\cite{ChenPKMY18} & Dilated-ResNet-$101$ & $70.4$ & $42.6$ & $86.4$ & $67.7$\\
			LC~\cite{LiLLLT17}& - & $71.1$ & - & - & - \\
			Piecewise~\cite{LinSHR16}& VGG-$16$ & $71.6$ & $51.7$ & $87.3$ & $74.1$\\
			FRRN~\cite{PohlenHML17}& - & $71.8$ & $45.5$ & $88.9$ & $75.1$\\
			RefineNet~\cite{LinMSR17}& ResNet-$101$ & $73.6$ & $47.2$ & $87.9$ & $70.6$\\
			PEARL~\cite{JinLXSLYCDLJFY17} & Dilated-ResNet-$101$ & $75.4$ & $51.6$ & $89.2$ & $75.1$ \\
			DSSPN~\cite{LiangZX18} & Dilated-ResNet-$101$ & $76.6$ & $56.2$ & $89.6$ & $77.8$\\
			LKM~\cite{PengZYLS17}& ResNet-$152$ & $76.9$ & - & - & - \\
			DUC-HDC~\cite{WangCYLHHC18}& - & $77.6$ & $53.6$ & $90.1$ & $75.2$\\
			SAC~\cite{ZhangTZLY17} & Dilated-ResNet-$101$ & $78.1$ & - & - & - \\
			DepthSeg~\cite{KongF18} & Dilated-ResNet-$101$ & $78.2$& - & - & - \\
			ResNet38~\cite{WuSH16e} & WResNet-38 &$78.4$ &$59.1$ &$90.9$ &$78.1$ \\
			BiSeNet~\cite{YuWPGYS182} & ResNet-$101$ & $78.9$ & - & - & - \\
			DFN~\cite{YuWPGYS18} & ResNet-$101$ & $79.3$ & - & - & - \\
			PSANet~\cite{ZhaoZLSLLJ18} & Dilated-ResNet-$101$ & $80.1$ & - & - & - \\
			PADNet~\cite{OWS18} & Dilated-ResNet-$101$ & $80.3$ & $58.8$ & $90.8$ & $78.5$\\
			DenseASPP~\cite{ZhaoSQWJ17} & WDenseNet-$161$ & $80.6$ & $59.1$ & $90.9$ & $78.1$ \\
			\hline
			Our approach &  HRNetV$2$-W$48$  & $\mathbf{81.6}$ & $\mathbf{61.8}$ & $\mathbf{92.1}$ & $\mathbf{82.2}$ \\
			\hline
		\end{tabular}
		\vspace{-3mm}
	\end{table}

Table~\ref{tab:cityscapevalresults} provides the comparison with several representative methods
on the Cityscapes validation set
in terms of parameter and computation complexity and mIoU class.
(i) HRNetV$2$-W$40$ ($40$ indicates the width of the high-resolution convolution), with similar model size to DeepLabv$3$+
and much lower computation complexity, gets better performance:
$4.7$ points gain over UNet++, $1.7$ points gain over DeepLabv3
and about $0.5$ points gain over PSPNet, DeepLabv3+.
(ii) HRNetV$2$-W$48$, with similar model size to PSPNet and much lower computation complexity, achieves much significant improvement:
$5.6$ points gain over UNet++, $2.6$ points gain over DeepLabv3
and about $1.4$ points gain over PSPNet, DeepLabv3+.
In the following comparisons,
we adopt HRNetV$2$-W$48$ that is pretrained on ImageNet
\footnote{The
description about
ImageNet pretraining
is given in the Appendix.}
and has similar model size
as most Dilated-ResNet-$101$ based methods.

Table~\ref{tab:cityscaperesults} provides the comparison of our method
with state-of-the-art methods on the Cityscapes test set.
All the results are with six scales and flipping.
Two cases w/o using coarse data are evaluated:
One is about the model learned
on the \texttt{train} set,
and the other is about the model
learned on the \texttt{train+valid} set.
In both cases,
HRNetV$2$-W$48$ achieves the best performance
and outperforms the previous state-of-the-art
by $1$ point.
	
	\renewcommand{\arraystretch}{1.3}
	\setlength{\tabcolsep}{3.0pt}
	\begin{table}[t]
	\scriptsize
	\centering
	\caption{Semantic segmentation results on PASCAL-context. The methods are
	evaluated on $59$ classes and $60$ classes.}
	\label{tab:pasctxresults}
	\begin{tabular}{l|l|cc}
		\hline\noalign{\smallskip}
		  & backbone & mIoU ($59$ classes) & mIoU ($60$ classes) \\
		\hline
		
		\hline
		FCN-$8$s~\cite{ShelhamerLD17} & VGG-$16$ & - & $35.1$ \\
		BoxSup~\cite{DaiHS15} & - & - & $40.5$ \\
		HO\_CRF~\cite{ArnabJ0T16} & - & - & $41.3$ \\
		Piecewise~\cite{LinSHR16} & VGG-$16$ & - &$43.3$ \\
		DeepLab-v$2$~\cite{ChenPKMY18} & Dilated-ResNet-$101$ & -& $45.7$ \\
		RefineNet~\cite{LinMSR17} & ResNet-$152$ & - & $47.3$ \\
		UNet++~\cite{ZhouSTL18} & ResNet-$101$ & $47.7$ & - \\
		PSPNet~\cite{ZhaoSQWJ17} & Dilated-ResNet-$101$ & $47.8$ & - \\
		Ding et al.~\cite{DingJSL018} & ResNet-$101$ & $51.6$ & - \\
		EncNet~\cite{0005DSZWTA18} & Dilated-ResNet-$101$ & $52.6$ & - \\
		\hline
		Our approach & HRNetV$2$-W$48$ &  $\mathbf{54.0}$ & $\mathbf{48.3}$ \\	
		\hline
	\end{tabular}
	\end{table}

\renewcommand{\arraystretch}{1.3}
	\setlength{\tabcolsep}{2.8pt}
	\begin{table}[t]
	\scriptsize
	\centering
	\caption{Semantic segmentation results on LIP. Our method doesn't exploit
	any extra information, e.g., pose or edge. }
	\label{tab:lipresults}
	\begin{tabular}{l|lc|ccc}
		\hline
		\noalign{\smallskip}
		 & backbone & extra. & pixel acc. & avg. acc. & mIoU \\
		\hline
		
		\hline
		Attention+SSL~\cite{GongLSL17} & VGG$16$ & Pose & $84.36$ & $54.94$ & $44.73$ \\
		DeepLabV$3$+~\cite{ChenZPSA18} & Dilated-ResNet-$101$ & - & $84.09$ & $55.62$ & $44.80$ \\
		MMAN~\cite{LuoZZGYY18} & Dilated-ResNet-$101$ & - & - & - & $46.81$ \\
		SS-NAN~\cite{ZhaoLNZCWFY17} & ResNet-$101$ & Pose & $87.59$ & $56.03$ & $47.92$ \\
		MuLA~\cite{NieFY18} & Hourglass & Pose & $88.50$ & $60.50$ & $49.30$ \\
		JPPNet~\cite{XL18} & Dilated-ResNet-$101$ & Pose & $86.39$ & $62.32$ & $51.37$ \\
		CE2P~\cite{TL18}  & Dilated-ResNet-$101$ & Edge & $87.37$ & $63.20$ & $53.10$ \\
		\hline
		Our approach & HRNetV$2$-W$48$ & N & $\mathbf{88.21}$ & $\mathbf{67.43}$ & $\mathbf{55.90}$ \\	
		\hline
	\end{tabular}
	 \vspace{-3mm}
	\end{table}
	
\vspace{.1cm}
\noindent\textbf{PASCAL context.}
The PASCAL context dataset~\cite{MottaghiCLCLFUY14}  includes $4,998$ scene images for training and
$5,105$ images for testing with $59$ semantic labels and $1$ background label.

The data augmentation and learning rate policy are the same as Cityscapes.
Following the widely-used training strategy~\cite{0005DSZWTA18, DingJSL018},
we resize the images to $480\times480$ and set the initial learning rate to $0.004$
and weight decay to $0.0001$. The batch size is $16$ and the number of iterations is $60K$.

We follow the standard testing procedure~\cite{0005DSZWTA18, DingJSL018}.
The image is resized to $480\times480$
and then fed into our network.
The resulting $480\times480$ label maps are then resized to the
original image size.
We evaluate the performance of our approach and other approaches
using six scales and flipping.

Table~\ref{tab:pasctxresults}
provides the comparison of our method
with state-of-the-art methods.
There are two kinds of evaluation schemes:
mIoU over $59$ classes and $60$ classes ($59$ classes + background).
In both cases,
HRNetV$2$-W$48$ performs superior to previous state-of-the-arts.

\vspace{.1cm}
\noindent\textbf{LIP.}
The LIP dataset \cite{GongLSL17}
contains $50,462$ elaborately annotated human images,
which are divided into $30,462$ training images,
and $10,000$ validation images. The methods are evaluated on
$20$ categories ($19$ human part labels and
$1$ background label).
Following the standard training and testing settings~\cite{TL18}, the images are resized to $473\times473$ and
the performance is evaluated
on the average of the segmentation maps of the original and flipped images.

The data augmentation and learning rate policy are the same as Cityscapes.
The training strategy follows the recent setting~\cite{TL18}.
We set the initial learning rate to $0.007$ and
the momentum to $0.9$ and the weight decay to $0.0005$.
The batch size is $40$ and the number of iterations is $110$K.

Table~\ref{tab:lipresults}
provides the comparison of our method
with state-of-the-art methods.
The overall performance of HRNetV$2$-W$48$
performs the best
with fewer parameters and lighter computation cost.
We also would like to mention that
our networks do not use extra information such as pose or edge.

\renewcommand{\arraystretch}{1.3}
	\begin{table}[t]
	    \centering\setlength{\tabcolsep}{2.9pt}
	    \scriptsize
	    \caption{GFLOPs and \#parameters
	    of Faster R-CNN for COCO object detection.
	    The numbers are obtained
	    with the input size $800 \times 1200$
	    and
	    $512$ proposals fed into R-CNN.
	    ResNet-$x$-FPN (R-$x$),
	    X-$101$-$64$$\times$$4$d (X-101),
	    HRNetV2p-W$x$ (H-$x$).
	    }
	    \label{tab:det_training_comparision}
	    \begin{tabular}{l|rr|rr|rr|rr}
	        \hline \noalign{\smallskip}
	        ~ & R-$50$ & H-$18$& R-$101$ &H-$32$& R-$152$ &H-$40$ & X-$101$ & H-$48$\\
	        \hline
	
	        \hline
	        \#param. (M) & $39.8$ & $26.2$ & $57.8$ & $45.0$ & $72.7$ & $60.5$ & $94.9$ & $79.4$ \\
	        GFLOPs & $172.3$ & $159.1$ & $239.4$ & $245.3$ & $306.4$ & $314.9$ & $381.8$ &  $399.1$\\
	        \hline
	    \end{tabular}
	\end{table}

\renewcommand{\arraystretch}{1.3}
	\begin{table}[t]
	\setlength{\tabcolsep}{4.4pt}
	\scriptsize
	\centering
	\caption{Object detection results evaluated on COCO \texttt{val}
	in the Faster R-CNN framework.
	LS = learning schedule.}
	\label{tab:object_detection_fpn}
	\begin{tabular}{l|c|ccc|ccc}
		\hline \noalign{\smallskip}
		backbone & LS & AP & AP$_{50}$ & AP$_{75}$ & AP$_S$ & AP$_M$ & AP$_L$\\
		\hline
		
		\hline

	  	ResNet-$50$-FPN  & $1\times$  & $36.7$ & $58.3$ & $39.9$ & $20.9$ & $39.8$ & $47.9$ \\
		HRNetV$2$p-W$18$ & $1\times$ & $36.2$ & $57.3$ & $39.3$ & $20.7$ & $39.0$ & $46.8$ \\
	    ResNet-$50$-FPN & $2\times$  & $37.6$ & $58.7$ & $41.3$ & $21.4$ & $\mathbf{40.8}$ & $\mathbf{49.7}$ \\
		HRNetV$2$p-W$18$ & $2\times$ & $\mathbf{38.0}$ & $\mathbf{58.9}$ & $\mathbf{41.5}$ & $\mathbf{22.6}$ & $\mathbf{40.8}$ & $49.6$ \\
	    \hline
	    ResNet-$101$-FPN & $1\times$ & $39.2$ & $61.1$ & $43.0$ & $22.3$ & $42.9$ & $50.9$ \\
		HRNetV$2$p-W$32$ & $1\times$ & $39.6$ & $61.0$ & $43.3$ & $23.7$ & $42.5$ & $50.5$ \\
		ResNet-$101$-FPN & $2\times$ & $39.8$ & $61.4$ & $43.4$ & $22.9$ & $43.6$ & $52.4$ \\
	    HRNetV$2$p-W$32$ & $2\times$ & $\mathbf{40.9}$ & $\mathbf{61.8}$ & $\mathbf{44.8}$ & $\mathbf{24.4}$ & $\mathbf{43.7}$ & $\mathbf{53.3}$  \\
	    \hline
	
	    ResNet-$152$-FPN & $1\times$ & $39.5$ & $61.2$ & $43.0$ & $22.1$ & $43.3$ & $51.8$ \\
	    HRNetV$2$p-W$40$ & $1\times$ & $40.4$ & $61.8$ & $44.1$ & $23.8$ & $43.8$ & $52.3$  \\

	    ResNet-$152$-FPN & $2\times$ & $40.6$ & $61.9$ & $44.5$ & $22.8$ & $44.0$ & $53.1$ \\

	    HRNetV$2$p-W$40$ & $2\times$ & $\mathbf{41.6}$ & $\mathbf{62.5}$ & $\mathbf{45.6}$ & $\mathbf{23.8}$ & $\mathbf{44.9}$ & $\mathbf{53.8}$  \\
	    \hline
	
	    X-$101$-$64$$\times$$4$d-FPN & $1\times$ & $41.3$ & $\mathbf{63.4}$ & $45.2$ & $24.5$ & $\mathbf{45.8}$ & $53.3$ \\
	    HRNetV$2$p-W$48$ & $1\times$ & $41.3$ & $62.8$ & $45.1$ & $\mathbf{25.1}$ & $44.5$ & $52.9$  \\
	    X-$101$-$64$$\times$$4$d-FPN & $2\times$ & $40.8$ & $62.1$ & $44.6$ & $23.2$ & $44.5$ & $53.7$ \\
	    HRNetV$2$p-W$48$ & $2\times$ & $\mathbf{41.8}$ & $62.8$ & $\mathbf{45.9}$ & $25.0$ & $44.7$ & $\mathbf{54.6}$  \\

	    \hline
	\end{tabular}
	\vspace{-1.5mm}
	\end{table}

\renewcommand{\arraystretch}{1.3}
	\begin{table}[t]
	\setlength{\tabcolsep}{2.3pt}
	\centering
	\caption{Object detection results evaluated on COCO \texttt{val}
	in the Mask R-CNN framework.
	LS = learning schedule.}
	\label{tab:object_detection_maskrcnn}
	\scriptsize
	\begin{tabular}{l|c|cccc|cccc}
		\hline \noalign{\smallskip}
		\multirow{2}{*}{backbone} &
		\multirow{2}{*}{LS} & \multicolumn{4}{c|}{mask} & \multicolumn{4}{c}{bbox} \\
		\cline{3-10}&  & AP & AP$_S$ & AP$_M$ & AP$_L$ & AP & AP$_S$ & AP$_M$ & AP$_L$ \\
		\hline

		\hline
	    ResNet-$50$-FPN & $1\times$  & $34.2$ & $15.7$ & $36.8$ & $50.2$ & $37.8$ & $22.1$ & $40.9$ & $49.3$  \\
		HRNetV$2$p-W$18$ & $1\times$ & $33.8$ & $15.6$ & $35.6$ & $49.8$ & $37.1$ & $21.9$ & $39.5$ & $47.9$  \\
	    ResNet-$50$-FPN & $2\times$  & $35.0$ & $16.0$ & $\mathbf{37.5}$ & $\mathbf{52.0}$ & $38.6$ & $21.7$ & $41.6$ & $50.9$  \\
		HRNetV$2$p-W$18$ & $2\times$ & $\mathbf{35.3}$ & $\mathbf{16.9}$ & $\mathbf{37.5}$ & $51.8$ & $\mathbf{39.2}$ & $\mathbf{23.7}$ & $\mathbf{41.7}$ & $\mathbf{51.0}$  \\
	    \hline
		ResNet-$101$-FPN & $1\times$ & $36.1$ & $16.2$ & $39.0$ & $53.0$ & $40.0$ & $22.6$ & $43.4$ & $52.3$  \\
		HRNetV$2$p-W$32$ & $1\times$ & $36.7$ & $17.3$ & $39.0$ & $53.0$ & $40.9$ & $24.5$ & $43.9$ & $52.2$  \\
		ResNet-$101$-FPN & $2\times$ & $36.7$ & $17.0$ & $39.5$ & $54.8$ & $41.0$ & $23.4$ & $44.4$ & $53.9$  \\
		HRNetV$2$p-W$32$ & $2\times$ & $\mathbf{37.6}$ & $\mathbf{17.8}$ & $\mathbf{40.0}$ & $\mathbf{55.0}$ & $\mathbf{42.3}$ & $\mathbf{25.0}$ & $\mathbf{45.4}$ & $\mathbf{54.9}$ \\
		\hline
	\end{tabular}
	\vspace{-0.4cm}
	\end{table}

\renewcommand{\arraystretch}{1.3}
    \begin{table*}[ht]
	\setlength{\tabcolsep}{9pt}
	\centering
	\caption{Comparison with the state-of-the-art single-model object detectors on COCO \texttt{test-dev} without mutli-scale training and testing.
	We obtain the results of Faster R-CNN and Cascade R-CNN
    by using our implementations publicly
	available from the mmdetection platform\cite{mmdetection2018}
	except that $^*$
	is from the original paper~\cite{CaiV18}.}
	\scriptsize
	\label{tab:recent_object_detection_results_single}
	\begin{tabular}{l|lcr|ccc|ccc}
    	\hline \noalign{\smallskip}
    	 & backbone & size & LS & AP & AP$_{50}$ & AP$_{75}$ & AP$_S$ & AP$_M$ & AP$_L$\\

		\hline
		
		\hline
		MLKP \cite{WangWGLZ18} & VGG$16$ & - & - &  $28.6$ & $52.4$ & $31.6$ & $10.8$ & $33.4$ & $45.1$ \\
		
		STDN \cite{ZhouNGHX18} & DenseNet-$169$ & $513$ & - & $31.8$ & $51.0$ & $33.6$ & $14.4$ & $36.1$ & $43.4$\\
		
		DES \cite{ZhangQX0WY18} & VGG$16$ & $512$  & - & $32.8$ & $53.2$ & $34.6$ & $13.9$ & $36.0$ & $47.6$ \\
		CoupleNet \cite{ZhuZWZWL17} & ResNet-$101$ & - & - & $33.1$ & $53.5$ & $35.4$ & $11.6$ & $36.3$ & $50.1$ \\
		DeNet \cite{Tychsen-SmithP17} & ResNet-$101$ & $512$ & - & $33.8$ & $53.4$ & $36.1$ & $12.3$ & $36.1$ & $50.8$ \\
	    RFBNet \cite{LiuHW18} & VGG$16$ & $512$ & -  & $34.4$ & $55.7$ &  $36.4$ & $17.6$ & $37.0$ & $47.6$ \\
		DFPR \cite{KongSHL18} & ResNet-$101$ & $512$ & $1\times$ & $34.6$ & $54.3$ & $37.3$ & - & - & - \\

	    PFPNet \cite{KimKSKK18} & VGG$16$ & $512$ & - & $35.2$ & $57.6$ &  $37.9$ & $18.7$ & $38.6$ & $45.9$ \\
	
	    RefineDet\cite{ZhangWBLL18} & ResNet-$101$ & $512$ & - & $36.4$ & $57.5$ & $39.5$ & $16.6$ & $39.9$ & $51.4$  \\

	    Relation Net \cite{HuGZDW18} & ResNet-$101$ & $600$ & - & $39.0$ & $58.6$ & $42.9$ & - & - & - \\
	    C-FRCNN \cite{ChenHT18} & ResNet-$101$ & $800$ & $1\times$ & $39.0$ & $59.7$ & $42.8$ & $19.4$ & $42.4$ & $53.0$ \\
	    RetinaNet \cite{LinGGHD17} & ResNet-$101$-FPN & $800$ & $1.5\times$ & $39.1$ & $59.1$ & $42.3$ & $21.8$ & $42.7$ & $50.2$ \\

        Deep Regionlets \cite{XuLWRBC18} & ResNet-$101$ & $800$ & $1.5\times$  & $39.3$ & $59.8$ &  - & $21.7$ & $43.7$ & $50.9$ \\
		
		FitnessNMS \cite{Tychsen-SmithP18} & ResNet-$101$ & $768$ & - & $39.5$ & $58.0$ & $42.6$ & $18.9$ & $43.5$ & $54.1$ \\
        DetNet \cite{LiPYZDS18} & DetNet$59$-FPN & $800$ & $2\times$ & $40.3$ & $62.1$ & $43.8$ & $23.6$ & $42.6$ & $50.0$ \\		
	    CornerNet \cite{LawD18} & Hourglass-$104$ & $511$ & - & $40.5$ & $56.5$ & $43.1$ & $19.4$ & $42.7$ & $53.9$ \\
	    M2Det \cite{M2DETQ} & VGG$16$ & $800$ & $\sim 10\times$ & $41.0$ & $59.7$ & $45.0$ & $22.1$ & $\mathbf{46.5}$ & $53.8$ \\
	
	    \hline
        Faster R-CNN \cite{LinDGHHB17} & ResNet-$101$-FPN & $800$ & $1\times$ & $39.3$ & $61.3$ & $42.7$ & $22.1$ & $42.1$ & $49.7$ \\
        Faster R-CNN & HRNetV$2$p-W$32$ & $800$ & $1\times$ & $39.5$ & $61.2$ & $43.0$ & $23.3$ & $41.7$ & $49.1$  \\
        \arrayrulecolor{gray}\hline\arrayrulecolor{black}
		Faster R-CNN \cite{LinDGHHB17} & ResNet-$101$-FPN & $800$ & $2\times$ & $40.3$ & $61.8$ & $43.9$ & $22.6$ & $43.1$ & $51.0$ \\
		Faster R-CNN & HRNetV$2$p-W$32$ & $800$ & $2\times$  & $41.1$ & $62.3$ & $44.9$ & $24.0$ & $43.1$ & $51.4$ \\	
		\arrayrulecolor{gray}\hline\arrayrulecolor{black}
		Faster R-CNN \cite{LinDGHHB17} & ResNet-$152$-FPN & $800$ & $2\times$ & $40.6$ & $62.1$ & $44.3$ & $22.6$ & $43.4$ & $52.0$ \\
		Faster R-CNN & HRNetV$2$p-W$40$ & $800$ & $2\times$  & $42.1$ & $63.2$ & $46.1$ & $24.6$ & $44.5$ & $52.6$ \\
		\arrayrulecolor{gray}\hline\arrayrulecolor{black}
	    Faster R-CNN \cite{mmdetection2018} & X-$101$-$64$$\times4$d-FPN & $800$ & $2\times$ & $41.1$ & $62.8$ & $44.8$ & $23.5$ & $44.1$ & $52.3$ \\
		Faster R-CNN & HRNetV$2$p-W$48$ &  $800$ & $2\times$ & $42.4$ & $\mathbf{63.6}$ & $46.4$ & $24.9$ & $44.6$ & $53.0$ \\

	   \arrayrulecolor{gray}\hline\arrayrulecolor{black}
	    Cascade R-CNN \cite{CaiV18}$^*$ & ResNet-$101$-FPN & $800$ & $\sim 1.6\times$ & $42.8$ & $62.1$ & $46.3$  & $23.7$ & $45.5$ & $55.2$ \\
	    Cascade R-CNN & ResNet-$101$-FPN & $800$ & $\sim 1.6\times$ & $43.1$ & $61.7$ & $46.7$ & $24.1$ & $45.9$ & $55.0$ \\
	    Cascade R-CNN & HRNetV$2$p-W$32$  & $800$ & $\sim 1.6\times$ & $\mathbf{43.7}$ & $62.0$ & $\mathbf{47.4}$ & $\mathbf{25.5}$ & $46.0$ & $\mathbf{55.3}$ \\
		\arrayrulecolor{black}\hline
	\end{tabular}
	\end{table*}

\subsection{COCO Object Detection}
We apply our multi-level representations (HRNetV$2$p)\footnote{Same as FPN~\cite{LinGGHD17},
we also use $5$ levels.},
shown in Figure~\ref{fig:highresolutionhead} (c),
in the Faster R-CNN~\cite{RenHG015} and Mask R-CNN~\cite{HeGDG17} frameworks.
We perform the evaluation on the MS-COCO $2017$ detection dataset,
which contains $\sim118$k images for training,
$5$k for validation (\texttt{val}) and $\sim20$k testing without provided annotations (\texttt{test-dev}).
The standard COCO-style evaluation is adopted.

We train the models for both our HRNetV$2$p and the ResNet
on the public mmdetection platform~\cite{mmdetection2018}
with the provided training setup,
except that we use the learning rate schedule suggested in~\cite{DBLP:journals/corr/abs-1811-08883}
for $2\times$.
The data is augmented
by standard horizontal flipping. The input images are resized such that the shorter edge is 800 pixels \cite{LinDGHHB17}.
Inference is performed on a single image scale.

Table~\ref{tab:det_training_comparision}
summarizes \#parameters and GFLOPs.
Table~\ref{tab:object_detection_fpn}
and Table~\ref{tab:object_detection_maskrcnn}
report the detection results on COCO \texttt{val}.
There are several observations.
(i) The model size and computation complexity of HRNetV$2$p-W$18$ (HRNetV$2$p-W$32$)
are smaller than ResNet-$50$-FPN (ResNet-$101$-FPN).
(ii) With $1\times$,
HRNetV2p-W$32$ performs better than ResNet-$101$-FPN.
HRNetV2p-W$18$ performs worse than ResNet-$50$-FPN,
which might come from insufficient optimization iterations.
(iii) With $2\times$, HRNetV2p-W$18$ and HRNetV2p-W$32$ perform better
than ResNet-$50$-FPN and ResNet-$101$-FPN, respectively.

 Table~\ref{tab:recent_object_detection_results_single}
reports the comparison
of our network to state-of-the-art single-model object detectors on COCO \texttt{test-dev} without
using multi-scale training and
multi-scale testing that are done in~\cite{LIUQQSJ18, QiLSJ18, LiCYD18, SinghND18,SinghD18,PengXLJZJYS18}.
In the Faster R-CNN framework,
our networks perform better than ResNets with similar
parameter and computation complexity:
HRNetV$2$p-W$32$ vs. ResNet-$101$-FPN,
HRNetV$2$p-W$40$ vs. ResNet-$152$-FPN,
HRNetV$2$p-W$48$ vs. X-$101$-$64\times4$d-FPN.
In the Cascade R-CNN framework,
our HRNetV$2$p-W$32$ performs better.

\subsection{Facial Landmark Detection}
Facial landmark detection
a.k.a. face alignment
is a problem of detecting the keypoints
from a face image.
We perform the evaluation over
four standard datasets:
WFLW~\cite{Wu0YWC018},
AFLW~\cite{KostingerWRB11},
COFW~\cite{Burgos-ArtizzuPD13},
and $300$W~\cite{SagonasTZP13}.
We mainly use the normalized mean error (NME) for evaluation.
We use the inter-ocular distance as normalization
for WFLW, COFW, and $300$W, and the face bounding box as normalization
for AFLW.
We also report area-under-the-curve scores (AUC) and failure rates.

We follow the standard scheme~\cite{Wu0YWC018} for training.
All the faces are cropped by the provided boxes according
to the center location and resized to $256 \times 256$.
We augment the data by $\pm30$ degrees in-plane rotation,
$0.75-1.25$ scaling, and
randomly flipping.
The base learning rate is $0.0001$ and is dropped to
$0.00001$ and $0.000001$ at the $30$th and $50$th epochs. The models
are trained for $60$ epochs with the batch size of $16$ on
one GPU. Different from semantic segmentation,
the heatmaps are not upsampled from $1/4$ to the input size,
and the loss function is optimized over
the $1/4$ maps.

At testing,
each keypoint location is predicted
by transforming the highest heatvalue location from $1/4$
to the original image space
and adjusting it with a quarter offset
in the direction from the highest response
to the second highest response~\cite{ChenSWLY17}.

We adopt HRNetV$2$-W$18$ for face landmark detection
whose parameter and computation cost
are similar to or smaller than
models with widely-used backbones: ResNet-$50$ and Hourglass~\cite{NewellYD16}.
HRNetV$2$-W$18$: \#parameters $=9.3$M, GFLOPs $=4.3$G;
ResNet-$50$: \#parameters $=25.0$M, GFLOPs $=3.8$G;
Hourglass: \#parameters $=25.1$M, GFLOPs $=19.1$G.
The numbers are obtained on the input size $256 \times 256$.
It should be noted that
the facial landmark detection methods adopting ResNet-$50$ and Hourglass as backbones
introduce extra parameter and computation overhead.

\vspace{.1cm}
\noindent\textbf{WFLW.}
The WFLW dataset \cite{Wu0YWC018} is a recently-built
dataset based on the WIDER Face \cite{YangLLT16}.
There are $7,500$ training and $2,500$ testing images
with $98$ manual annotated landmarks.
We report the results on the test set and several subsets:
large pose ($326$ images), expression ($314$ images),
illumination ($698$ images), make-up ($206$ images), occlusion ($736$ images) and blur ($773$
images).

Table \ref{table:comparison_wflw_testset} provides the comparison of our method with state-of-the-art methods.
Our approach is significantly better than other methods on the test set and all the subsets,
including LAB that exploits extra boundary information~\cite{Wu0YWC018} and PDB that uses stronger data augmentation~\cite{FengKA0W18}.

\renewcommand{\arraystretch}{1.3}
\begin{table}[t]
\scriptsize
\setlength{\tabcolsep}{0.5pt}
\centering
\caption{Facial landmark detection results
(NME) on WFLW \texttt{test} and $6$ subsets:
pose,
expression (expr.),
illumination (illu.),
make-up (mu.),
occlusion (occu.)
and blur.
LAB~\cite{Wu0YWC018} is trained with extra boundary information (B).
PDB~\cite{FengKA0W18} adopts stronger data augmentation (DA).
Lower is better.}
\label{table:comparison_wflw_testset}
\begin{tabular}{l|l|r|r|r|r|r|r|r }
\hline \noalign{\smallskip}
   & backbone & test & pose & expr. & illu. & mu & occu. & blur\\
\hline

\hline
 ESR \cite{CaoWWS12}& - & $11.13$ & $25.88$ & $11.47$ & $10.49$ & $11.05$ & $13.75$ & $12.20$\\
SDM \cite{XiongT13}& - &$10.29$ & $24.10$ & $11.45$ & $9.32$ & $9.38$ & $13.03$ & $11.28$\\
CFSS \cite{ZhuLLT15}& - &$9.07$ & $21.36$ & $10.09$ & $8.30$ & $8.74$ & $11.76$ & $9.96$\\
DVLN \cite{WuY17} & VGG-16&$6.08$ & $11.54$ & $6.78$ & $5.73$ & $5.98$ & $7.33$ & $6.88$\\
\hline
Our approach & HRNetV$2$-W$18$ & $\mathbf{4.60}$ & $\mathbf{7.94}$ & $\mathbf{4.85}$ & $\mathbf{4.55}$ & $\mathbf{4.29}$ & $\mathbf{5.44}$ & $\mathbf{5.42}$\\
\hline

\hline
\multicolumn{3}{l}{
\emph {Model trained with \texttt{extra} info.}}\\
\hline
LAB (w/ B)~\cite{Wu0YWC018}& Hourglass & $5.27$ & $10.24$ & $5.51$ & $5.23$ & $5.15$ & $6.79$ & $6.32$\\
PDB (w/ DA)~\cite{FengKA0W18}& ResNet-$50$ & $5.11$ & $8.75$ & $5.36$ & $4.93$ & $5.41$ & $6.37$ & $5.81$\\
\hline
\end{tabular}
\end{table}

\vspace{.1cm}
\noindent\textbf{AFLW.}
The AFLW \cite{KostingerWRB11} dataset is
a widely used benchmark dataset,
where each image has $19$ facial landmarks. Following \cite{ZhuLLT15, Wu0YWC018}, we train our models on $20,000$ training images, and report the results on the AFLW-Full set ($4,386$ testing images) and the AFLW-Frontal set ($1314$ testing images selected from $4386$ testing images).

Table \ref{table:comparison_aflw_testset} provides the comparison of our method with state-of-the-art methods.
Our approach achieves the best performance among methods without extra information and stronger data augmentation and even outperforms DCFE with extra $3$D information.
Our approach performs slightly worse than
LAB that uses extra boundary information~\cite{Wu0YWC018}
and PDB~\cite{FengKA0W18}
that uses stronger data augmentation.

\renewcommand{\arraystretch}{1.3}
\begin{table}[t]
\setlength{\tabcolsep}{12.5pt}
\scriptsize
\centering
\caption{Facial landmark detection results (NME) on AFLW.
DCFE~\cite{ValleBVB18} uses extra $3$D information ($3$D). Lower is better.}
\label{table:comparison_aflw_testset}
\begin{tabular}{l|l|c|c }
\hline \noalign{\smallskip}
 & backbone &  full &  frontal \\
\hline

\hline
RCN \cite{HonariYVP16} &- & $5.60$ & $5.36$\\
CDM \cite{YuHZYM13} & -& $5.43$ & $3.77$\\
ERT \cite{KazemiS14}&- & $4.35$ & $2.75$\\
LBF \cite{RenCW014} & -& $4.25$ & $2.74$\\
SDM \cite{XiongT13} &- & $4.05$ & $2.94$\\
CFSS \cite{ZhuLLT15}&- & $3.92$ & $2.68$ \\
RCPR \cite{Burgos-ArtizzuPD13}&- &$3.73$ & $2.87$\\
CCL \cite{ZhuLLT16}&- &$2.72$ & $2.17$\\
DAC-CSR \cite{FengKC0W17}& &$2.27$ & $1.81$\\
TSR \cite{LvSXCZ17}&VGG-S &$2.17$ & - \\
CPM + SBR \cite{DongYWW0S18}& CPM &$2.14$&-\\
SAN \cite{DongYO018}& ResNet-$152$ &$1.91$ & $1.85$\\
DSRN \cite{MiaoZLDAH18}& - &$1.86$ & -\\
LAB (w/o B) \cite{Wu0YWC018}& Hourglass & $1.85$ & $1.62$\\
\hline
Our approach & HRNetV2-W$18$ & $\mathbf{1.57}$ & $\mathbf{1.46}$ \\
\hline

\hline
\multicolumn{3}{l}{
\emph {Model trained with \texttt{extra} info.}}\\
\hline
DCFE (w/ $3$D)~\cite{ValleBVB18}& - &$2.17$ & - \\
PDB (w/ DA)~\cite{FengKA0W18}& ResNet-$50$&$1.47$ & -\\
LAB (w/ B)~\cite{Wu0YWC018}& Hourglass &$1.25$ & $1.14$\\
\hline
\end{tabular}
\end{table}

\vspace{.1cm}
\noindent\textbf{COFW.}
The COFW dataset \cite{Burgos-ArtizzuPD13}
consists of $1,345$ training
and $507$ testing faces
with occlusions,
where each image has $29$ facial landmarks.

Table~\ref{table:comparison_cofw_testset} provides the comparison of our method with state-of-the-art methods.
HRNetV$2$ outperforms other methods by a large margin.
In particular, it achieves the better performance than LAB with
extra boundary information and PDB with stronger data augmentation.

\renewcommand{\arraystretch}{1.3}
\begin{table}[t]
\setlength{\tabcolsep}{12.8pt}
\scriptsize
\centering
\caption{Facial landmark detection results on COFW \texttt{test}.
The failure rate is calculated at the threshold $0.1$.
Lower is better for NME and FR$_{0.1}$.
}
\begin{tabular}{l|l|cc }
\hline\noalign{\smallskip}
 & backbone & NME & FR$_{0.1}$\\
\hline

\hline
Human & - & $5.60$ & -\\
ESR \cite{CaoWWS12} &-&$11.20$&$36.00$\\
RCPR \cite{Burgos-ArtizzuPD13}& - &$8.50$ & $20.00$\\
HPM \cite{GhiasiF14}& - &$7.50$ & $13.00$ \\
CCR \cite{FengHKCW15}& -& $7.03$ & $10.90$ \\
DRDA \cite{ZhangKSC16}& - &$6.46$ & $6.00$ \\
RAR \cite{XiaoFXLYK16}& - &$6.03$ & $4.14$ \\
DAC-CSR \cite{FengKC0W17}& - &$6.03$ & $4.73$ \\
LAB (w/o B) \cite{Wu0YWC018}& Hourglass & $5.58$ & $2.76$\\
\hline
Our approach & HRNetV$2$-W$18$ & $\mathbf{3.45}$ & $\mathbf{0.19}$ \\
\hline

\hline
\multicolumn{3}{l}{
\emph {Model trained with \texttt{extra} info.}}\\
\hline
PDB (w/ DA)~\cite{FengKA0W18}& ResNet-$50$ & $5.07$ & $3.16$\\
LAB (w/ B)~\cite{Wu0YWC018}& Hourglass &$3.92$ & $0.39$ \\
\hline
\end{tabular}
\label{table:comparison_cofw_testset}
\end{table}

\vspace{.1cm}
\noindent\textbf{$300$W.}
The dataset~\cite{SagonasTZP13} is a combination
of HELEN~\cite{LeBLBH12}, LFPW ~\cite{BelhumeurJKK13}, AFW~\cite{ZhuR12}, XM2VTS~ and IBUG datasets,
where each face has $68$ landmarks.
Following~\cite{RenCWS16}, we use the $3,148$ training images, which
contains the training subsets of HELEN and LFPW and the full set of AFW.
We evaluate the performance
using two protocols, full set and test set.
The full set contains $689$ images and is further divided
into a common subset ($554$ images) from HELEN and LFPW, and a challenging subset
($135$ images) from IBUG.
The official test set, used for competition,  contains $600$ images ($300$ indoor and $300$ outdoor images).

Table \ref{table:comparison_300w_fullset}
provides the results
on the full set, and its two subsets:
common and challenging.
Table \ref{table:comparison_300w_testset} provides the results
on the test set.
In comparison to Chen et al. \cite{ChenSWLY17}
that uses Hourglass with large parameter and computation complexity
as the backbone,
our scores are better except the AUC$_{0.08}$ scores.
Our HRNetV$2$ gets the overall
best performance among methods without extra information and stronger data augmentation, and is even better than LAB with extra boundary information and DCFE~\cite{ValleBVB18} that explores extra $3$D information.

\renewcommand{\arraystretch}{1.3}
\begin{table}[t]
\setlength{\tabcolsep}{6.2pt}
\centering
\scriptsize
\caption{Facial landmark detection results (NME)
on $300$W:
common, challenging and full.
Lower is better.}

\label{table:comparison_300w_fullset}
\begin{tabular}{ l|l|ccc  }
\hline\noalign{\smallskip}
  & backbone &common & challenging & full \\
\hline

\hline
RCN \cite{HonariYVP16} &-& $4.67$ & $8.44$ & $5.41$\\
DSRN \cite{MiaoZLDAH18} &-& $4.12$ & $9.68$ & $5.21$ \\
PCD-CNN \cite{KumarC18} &-& $3.67$ & $7.62$ & $4.44$ \\
CPM + SBR \cite{DongYWW0S18} & CPM & $3.28$ & $7.58$ & $4.10$ \\
SAN \cite{DongYO018} &ResNet-152& $3.34$ & $6.60$ & $3.98$ \\
DAN \cite{KowalskiNT17} &-& $3.19$ & $5.24$ & $3.59$ \\
\hline
Our approach & HRNetV$2$-W$18$ &$\mathbf{2.87}$ & $\mathbf{5.15}$ & $\mathbf{3.32}$\\
\hline

\hline
\multicolumn{3}{l}{
\emph {Model trained with \texttt{extra} info.}}\\
\hline
LAB (w/ B) \cite{Wu0YWC018}& Hourglass & $2.98$ & $5.19$ & $3.49$ \\
DCFE (w/ $3$D) \cite{ValleBVB18}& - &$2.76$ & $5.22$ & $3.24$ \\
\hline
\end{tabular}
\end{table}

\renewcommand{\arraystretch}{1.3}
\begin{table}[t]
\setlength{\tabcolsep}{1.8pt}
\scriptsize
\centering
\caption{Facial landmark detection results on $300$W \texttt{test}.
DCFE~\cite{ValleBVB18} uses extra 3D information (3D).
LAB~\cite{Wu0YWC018} is trained with extra boundary information (B).
Lower is better for NME, FR$_{0.08}$ and FR$_{0.1}$,
and higher is better for AUC$_{0.08}$ and AUC$_{0.1}$.}
\label{table:comparison_300w_testset}
\begin{tabular}{l|l|c|c|c|c|c}
\hline\noalign{\smallskip}
 & backbone &NME & AUC$_{0.08}$ & AUC$_{0.1}$ & FR$_{0.08}$ & FR$_{0.1}$\\
\hline

\hline
Balt. et al. \cite{Baltrusaitis0M13} & - & - &$19.55$ &-& $38.83$ & -\\
ESR \cite{CaoWWS12}& - &$8.47$ & $26.09$ & - & $30.50$ & - \\
ERT \cite{KazemiS14}& - &$8.41$ & $27.01$ & - & $28.83$ & - \\
LBF \cite{RenCW014}& - &$8.57$ & $25.27$ & - & $33.67$ & - \\
Face++ \cite{ZhouFCJY13} & - & - &$32.81$ & - & $13.00$ & - \\
SDM \cite{XiongT13}& - &$5.83$ & $36.27$ & - & $13.00$ & - \\
CFAN \cite{ZhangSKC14}& - &$5.78$ & $34.78$ & - & $14.00$ & - \\
Yan et al. \cite{YanLYL13} & -&- &$34.97$&-&$12.67$&-\\
CFSS \cite{ZhuLLT15}& - &$5.74$ & $36.58$ & - & $12.33$ & - \\
MDM \cite{TrigeorgisSNAZ16}& - &$4.78$ & $45.32$ & - & $6.80$ & - \\
DAN \cite{KowalskiNT17}& - &$4.30$ & $47.00$ & - & $2.67$ & - \\
Chen et al. \cite{ChenSWLY17}& Hourglass &$3.96$ & $\mathbf{53.64}$ & - & $2.50$ & - \\
Deng et al. \cite{DengLYT16}& - & - & - & $47.52$ & - & $5.50$ \\
Fan et al. \cite{FanZ16}& - &- & - & $48.02$ & - & $14.83$ \\
DReg + MDM \cite{GulerTASZK17}& ResNet101 &- & - & $52.19$ & - & $3.67$ \\
JMFA \cite{DengTZZ17}&Hourglass &- & - & $54.85$ & - & $1.00$ \\
\hline
Our approach & HRNetV$2$-W$18$& $\mathbf{3.85}$  & 52.09  & $\mathbf{61.55}$ & $\mathbf{1.00}$ & $\mathbf{0.33}$  \\
\hline

\hline
\multicolumn{3}{l}{
\emph {Model trained with \texttt{extra} info.}}\\
\hline
LAB (w/ B) \cite{Wu0YWC018}& Hourglass &- & -& $58.85$ & - & $0.83$ \\
DCFE (w/ $3$D) \cite{ValleBVB18}& - &$3.88$ & $52.42$ & - & $1.83$ & - \\
\hline
\end{tabular}
\end{table}

\subsection{Empirical Analysis}
\label{sec:empiricalstudy}
\begin{figure}[t]
    \centering
    \footnotesize
    (a)~\includegraphics[height=4.7cm]{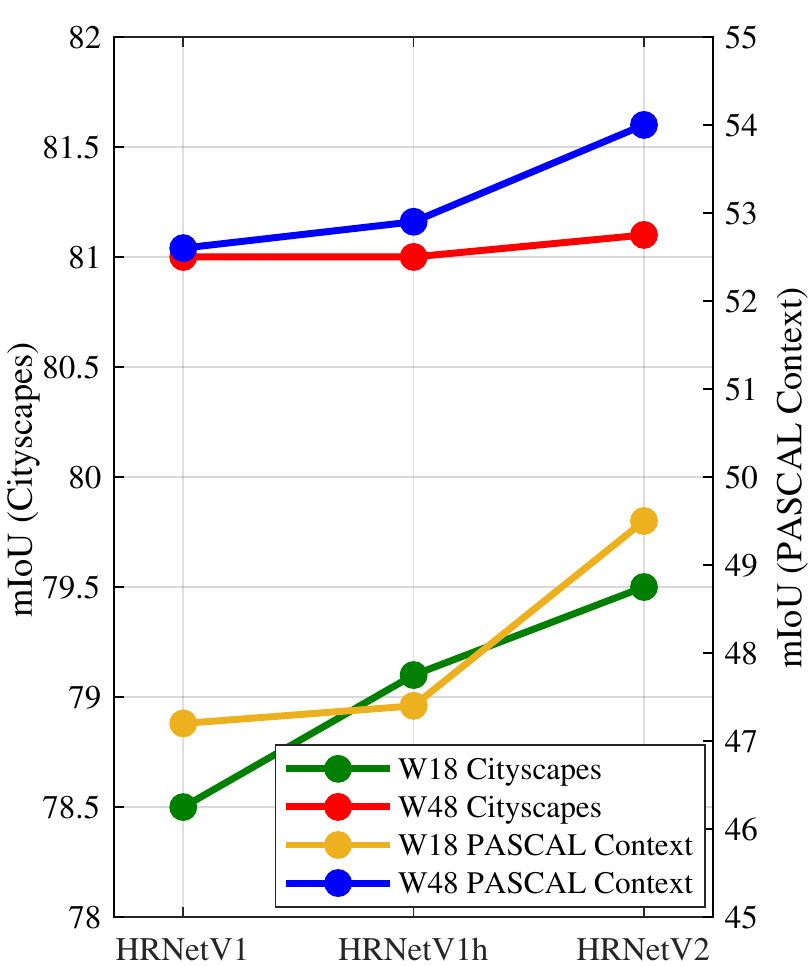}
    (b)~\includegraphics[height=4.7cm]{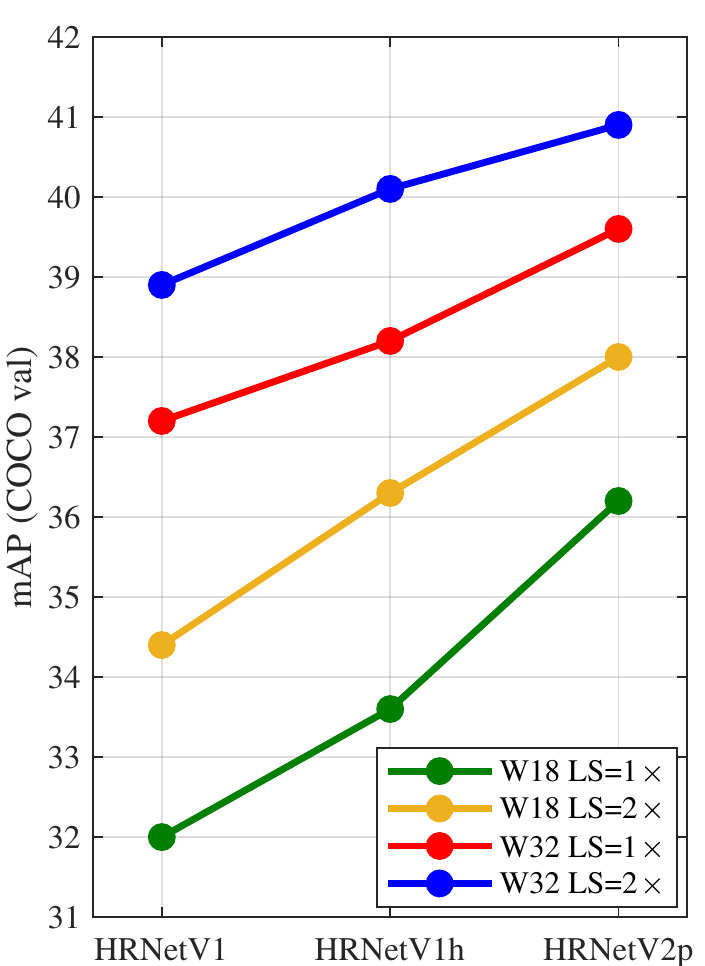}
    \caption{Empirical analysis.
    (a) Segmentation on Cityscapes \texttt{val} and PASCAL-Context \texttt{test} for comparing HRNetV$1$ and its variant HRNetV$1$h, and HRNetV$2$ (single scale and no flipping).
    (b) Object detection on COCO \texttt{val} for comparing HRNetV$1$ and its variant HRNetV$1$h, and HRNetV$2$p (LS = learning schedule).}
    \label{fig:empiricalstudy}
    \vspace{-0.3cm}
\end{figure}

We compare the modified networks, HRNetV$2$ and HRNetV$2$p,
to the original network~\cite{SunXLW19} (shortened as HRNetV$1$)
on semantic segmentation and COCO object detection.
The segmentation and object detection results, given in
Figure~\ref{fig:empiricalstudy} (a)
and Figure~\ref{fig:empiricalstudy} (b),
imply that HRNetV$2$ outperforms HRNetV$1$ significantly,
except that
the gain is minor
in the large model case
in segmentation for Cityscapes.
We also test a variant
(denoted by HRNetV$1$h),
which is built by appending a $1\times 1$ convolution to
increase the dimension of the output high-resolution representation.
The results in Figure~\ref{fig:empiricalstudy} (a) and Figure~\ref{fig:empiricalstudy} (b)
show that the variant achieves slight improvement
to HRNetV$1$,
implying that aggregating the representations
from low-resolution parallel convolutions in our HRNetV$2$
is essential for increasing the capability.

\section{Conclusions}
In this paper,
we empirically study
the high-resolution representation network
in a broad range of vision applications
with introducing a simple modification.
Experimental results demonstrate
the effectiveness of strong high-resolution representations
and multi-level representations
learned by the modified networks
on semantic segmentation, facial landmark detection
as well as object detection.
The project page is \url{https://jingdongwang2017.github.io/Projects/HRNet/}.

\section*{Appendix: Network Pretraining}
\begin{figure}
\footnotesize
    \centering
    \includegraphics[width=0.8\linewidth]{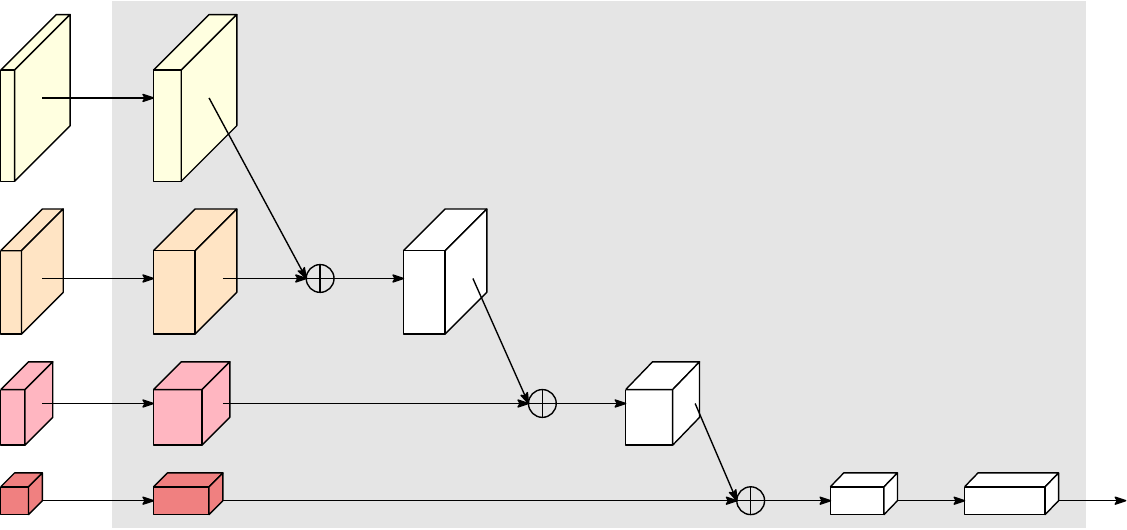}
    \caption{Representation for ImageNet classification.
    The input of the box
    is the representations
    of four resolutions.}
    \label{fig:classificationhead}
    \vspace{-4mm}
\end{figure}

We pretrain our network,
which is augmented by a classification head shown in Figure~\ref{fig:classificationhead},
on ImageNet~\cite{RussakovskyDSKS15}.
The classification head is described as below.
First, the four-resolution feature maps are fed into
a bottleneck and the output channels are
increased
from
$C$, $2C$, $4C$, and $8C$
to $128$, $256$, $512$, and $1024$, respectively.
Then, we downsample the high-resolution representation
by a $2$-strided $3 \times 3$ convolution
outputting $256$ channels
and add it to the representation of the second-high-resolution.
This process is repeated two times
to get $1024$ feature channels over the small resolution.
Last, we transform the $1024$ channels to $2048$ channels
through a $1 \times 1$ convolution, followed by a global average pooling operation.
The output $2048$-dimensional representation is fed into the classifier.

We adopt the same data augmentation scheme for training images as in \cite{HeZRS16},
and train our models for $100$ epochs with a batch size of $256$.
The initial learning rate is set to $0.1$
and is reduced by $10$ times at epoch $30$, $60$ and $90$.
We use SGD with a weight decay of $0.0001$ and a Nesterov momentum of $0.9$. We adopt standard single-crop testing, so that $224\times 224$ pixels are cropped from each image. The top-$1$
and top-$5$ error are reported on the validation set.

Table~\ref{tab:ImageNetClassificationComparison}
shows our ImageNet classification results.
As a comparison, we also report the results of ResNets.
We consider two types of residual units:
One is formed by a bottleneck,
and the other is formed by two $3 \times3 $ convolutions. We follow the
PyTorch implementation of ResNets and replace the $7 \times 7$
convolution in the input stem with two
$2$-strided $3 \times 3$ convolutions decreasing the resolution to $1/4$
as in our networks. When the residual units are formed by two $3 \times3 $ convolutions,
an extra bottleneck is used to increase the dimension of output feature maps from $512$ to $2048$.
One can see that under similar \#parameters
and GFLOPs, our results are comparable to and slightly better
than ResNets.

In addition, we look at the results
of two alternative schemes:
(i) the feature maps on each resolution go through a global pooling separately and then are concatenated
together to output a $15C$-dimensional representation vector,
named HRNet-W$x$-Ci;
(ii) the feature maps on each resolution are fed into several $2$-strided residual units (bottleneck, each dimension is increased to the double)
to increase the dimension to $512$,
and concatenate and average-pool them together to reach a $2048$-dimensional representation vector,
named HRNet-W$x$-Cii,
which is used in~\cite{SunXLW19}.
Table~\ref{tab:ImageNetClassificationAblationStudy}
shows such an ablation study.
One can see that the proposed manner is superior to the two alternatives.

	\begin{table}[ht]
	\setlength{\tabcolsep}{9.8pt}
	\scriptsize
	\centering
	\caption{ImageNet Classification results of HRNet and ResNets. The proposed method is named HRNet-W$x$-C.}
	\begin{tabular}{l|cc|cc}
		\hline
		& \#Params. & GFLOPs & top-1 err. & top-5 err. \\
		\hline
		
		\hline
		\multicolumn{5}{l}{\emph {Residual branch formed by two $3\times 3$ convolutions}}\\
		\hline
		ResNet-$38$ & $28.3$M & $3.80$ & $24.6\%$ & $7.4\%$ \\
		HRNet-W$18$-C& $21.3$M & $3.99$ & $\mathbf{23.1\%}$ & $\mathbf{6.5\%}$ \\
		\hline
		ResNet-$72$ & $48.4$M & $7.46$ & $23.3\%$ & $6.7\%$ \\
		HRNet-W$30$-C& $37.7$M & $7.55$ & $\mathbf{21.9\%}$ & $\mathbf{5.9\%}$ \\
		\hline
		ResNet-$106$& $64.9$M & $11.1$ & $22.7\%$ & $6.4\%$ \\	
		HRNet-W$40$-C& $57.6$M & $11.8$ & $\mathbf{21.1\%}$ & $\mathbf{5.6\%}$ \\
		\hline
		
		\hline
		\multicolumn{5}{l}{\emph {Residual branch formed by a bottleneck}}\\
		\hline
		ResNet-$50$& $25.6$M & $3.82$ & $23.3\%$ & $6.6\%$ \\
		HRNet-W$44$-C& $21.9$M & $3.90$ & $\mathbf{23.0\%}$ & $\mathbf{6.5\%}$ \\
		\hline
		ResNet-$101$& $44.6$M & $7.30$ & $21.6\%$ & $\mathbf{5.8\%}$ \\
		HRNet-W$76$-C& $40.8$M & $7.30$ & $\mathbf{21.5\%}$ & $\mathbf{5.8\%}$ \\
		\hline
		ResNet-$152$& $60.2$M & $10.7$ & $21.2\%$ & $5.7\%$ \\
		HRNet-W$96$-C& $57.5$M & $10.2$ & $\mathbf{21.0\%}$ & $\mathbf{5.6\%}$ \\
		\hline
	\end{tabular}
	\label{tab:ImageNetClassificationComparison}
	\vspace{-4mm}
	\end{table}

	\begin{table}[ht]
	\setlength{\tabcolsep}{8.8pt}
	\tiny
	\scriptsize
	\centering
	\caption{Ablation study on ImageNet classification
	by comparing our approach (abbreviated as HRNet-W$x$-C) with two alternatives: HRNet-W$x$-Ci and HRNet-W$x$-Cii (residual branch formed by two $3\times 3$ convolutions).}
	\begin{tabular}{l|cc|cc}
		\hline
		& \#Params. & GFLOPs & top-1 err. & top-5 err.\\
		\hline
	    HRNet-W$27$-Ci& $21.4$M & $5.55$ & $26.0\%$ & $7.7\%$ \\
        HRNet-W$25$-Cii& $21.7$M & $5.04$ & $24.1\%$ & $7.1\%$ \\
	    HRNet-W$18$-C& $21.3$M & $3.99$ & $\mathbf{23.1}\%$ & $\mathbf{6.5\%}$ \\
        \hline
        HRNet-W$36$-Ci& $37.5$M & $9.00$ & $24.3\%$ & $7.3\%$ \\
        HRNet-W$34$-Cii& $36.7$M & $8.29$ & $22.8\%$ & $6.3\%$ \\
	    HRNet-W$30$-C& $37.7$M & $7.55$ & $\mathbf{21.9\%}$ & $\mathbf{5.9\%}$ \\
        \hline
        HRNet-W$45$-Ci& $58.2$M & $13.4$ & $23.6\%$ & $7.0\%$ \\
        HRNet-W$43$-Cii& $56.3$ M & $12.5$ & $22.2\%$ & $6.1\%$ \\
	    HRNet-W$40$-C& $57.6$M & $11.8$ & $\mathbf{21.1\%}$ & $\mathbf{5.6\%}$ \\
	    \hline
	\end{tabular}
	\label{tab:ImageNetClassificationAblationStudy}
	\vspace{-5mm}
	\end{table}

\end{document}